\documentclass[5p,times]{elsarticle}
\usepackage{amsmath}
\usepackage{float} 
\usepackage{url}

\usepackage{amssymb}

\usepackage[figuresright]{rotating}

\usepackage[bookmarks=false]{hyperref}
    \hypersetup{colorlinks,
      linkcolor=blue,
      citecolor=blue,
      urlcolor=blue}

\date{}

\begin{document}
\begin{frontmatter}




\title{FDM-Bench: A Comprehensive Benchmark for Evaluating Large Language Models in Additive Manufacturing Tasks}


\author[a,b]{Ahmadreza Eslaminia}
\author[c]{Adrian Jackson}
\author[b]{Beitong Tian}
\author[c]{Avi Stern}
\author[c]{Hallie Gordon}
\author[c]{Rajiv Malhotra}
\author[b]{Klara Nahrstedt}
\author[d,a]{Chenhui Shao\corref{cor1}}
\ead{chshao@umich.edu}

\address[a]{Department of Mechanical Science and Engineering, University of Illinois at Urbana-Champaign, Urbana, IL 61801, USA}
\address[b]{Coordinated Science Laboratory, University of Illinois at Urbana-Champaign, Urbana, IL 61801, USA}
\address[c]{Department of Mechanical and Aerospace Engineering, Rutgers University, Piscataway, NJ 08854, USA}
\address[d]{Department of Mechanical Engineering, University of Michigan, Ann Arbor, MI 48109, USA}

\begin{abstract}
Fused Deposition Modeling (FDM) is a widely used additive manufacturing (AM) technique valued for its flexibility and cost-efficiency, with applications in a variety of industries including 
healthcare and aerospace. Recent technological developments have made affordable FDM machines accessible and encouraged adoption among diverse users. However, the design, planning, and production process in FDM require specialized interdisciplinary knowledge. Managing the complex parameters and resolving print defects in FDM remain challenging. These technical complexities form the most critical barrier preventing individuals without technical backgrounds and even professional engineers without training in other domains from participating in AM design and manufacturing. Large Language Models (LLMs), with their advanced capabilities in text and code processing, offer the potential for addressing these challenges in FDM. However, existing research on LLM applications in this field is limited, typically focusing on specific use cases without providing comprehensive evaluations across multiple models and tasks. To this end, we introduce FDM-Bench, a benchmark dataset designed to evaluate LLMs on FDM-specific tasks. FDM-Bench enables a thorough assessment by including user queries across various experience levels and G-code samples that represent a range of anomalies. We evaluate two closed-source models (GPT-4o and Claude 3.5 Sonnet) and two open-source models (Llama-3.1-70B and Llama-3.1-405B) on FDM-Bench. A panel of FDM experts assess the models’ responses to user queries in detail. Results indicate that closed-source models generally outperform open-source models in G-code anomaly detection, whereas Llama-3.1-405B demonstrates a slight advantage over other models in responding to user queries. These findings underscore FDM-Bench’s potential as a foundational tool for advancing research on LLM capabilities in FDM.
\end{abstract}

\cortext[cor1]{Corresponding author}

\end{frontmatter}


\section{Introduction}
\label{sec:intro}
Fused Deposition Modeling (FDM) is a widely adopted additive manufacturing (AM) technique valued for its versatility and cost-effectiveness~\cite{Zisopol}. Its ability to reduce lead times and accelerate prototyping cycles makes FDM particularly attractive to industries that prioritize rapid iteration~\cite{Sing, Aabith}. Therefore, this method is extensively used across industries, including manufacturing~\cite{Sing, Cleeman}, healthcare~\cite{BujCorral, Cailleaux}, aerospace~\cite{Kalender}, and consumer goods~\cite{Jeong, Tsegay, Jahangir}, for producing complex geometries and customized components. The accessibility of FDM has extended its reach beyond industrial applications, making it available to small businesses~\cite{Laplume}, research labs~\cite{Pearce}, and individuals.

Despite its broad applications, FDM is susceptible to quality issues such as low dimensional accuracy~\cite{HaghshenasGorgani} and poor surface quality~\cite{Dey}. Various process parameters, such as layer height, bed temperature, print speed, and fill angle, affect the quality of FDM-printed parts~\cite{Cleeman2023, Solomon, Zharylkassyn, Maurya}. These parameters interact in complex ways~\cite{Ajjarapu}, and their optimal configuration depends on numerous factors, such as machine type, filament materials, and nozzle dimensions~\cite{Hira, Lei}. Additionally, a wide range of print defects in FDM can stem from root causes related to both hardware and material properties~\cite{BaechleClayton}. Mitigating these defects requires careful diagnosis by knowledgeable operators~\cite{Kantaros}. Consequently, the challenges of parameter tuning, defect diagnosis, and defect mitigation create a steep learning curve that prevents the general public from designing their personalized parts and fabricating with FDM~\cite{Kantaros}. This highlights the need for robust solutions and standardized practices to plan and optimize production as well as streamline troubleshooting within FDM~\cite{HsiangLoh}.{\vadjust{\pagebreak}}

Large Language Models (LLMs) are advanced transformer-based models trained on extensive datasets. These models demonstrate promising capabilities across various natural language processing tasks, including interpreting and generating text and code~\cite{Ni}. Beyond these core functions, LLMs exhibit reasoning abilities that enable them to draw conclusions and address complex problems in novel domains~\cite{Devunuri, Kevian}. This adaptability has led to their increasing application across diverse fields, from customer support to data-driven problem-solving. These successful applications suggest the potential of LLMs for addressing challenges specific to FDM.

A few preliminary studies have explored the potential of LLMs in the FDM domain, particularly for answering questions and conducting G-code-related analyses. For example, Sriwastwa et al.~\cite{Sriwastwa} investigated whether ChatGPT, an LLM developed by OpenAI, can effectively respond to FDM-related questions posed by biomedical students. However, this study lacks quantitative evaluations and focuses solely on biomedical users. In G-code-related research, Badini et al.~\cite{Badini} assessed ChatGPT's capability to optimize FDM printing parameters based on inputted G-code. Although ChatGPT was reported to be effective in optimizing parameters, this study did not observe concrete, measurable improvements in print quality.

Another limitation of these studies is that they focus exclusively on ChatGPT, which represents only one of the state-of-the-art LLMs. To address this, Jignasu et al.~\cite{Jignasu} examined the capabilities of multiple open-source and closed-source models in manipulating G-codes for tasks such as part rotation and translation. However, this study also lacks a quantitative comparison between models. Despite some advancements in using LLMs for G-code interaction, a critical gap remains in applying these models for G-code-based anomaly detection. Accurately identifying defects requires that LLMs understand the syntax of G-code and can extract relevant printing parameters. Additionally, the models need to understand the relationships between these parameters and potential errors to provide effective detections.

Given these gaps, this study aims to evaluate the effectiveness of LLMs on a range of FDM-specific tasks by establishing a comprehensive benchmark dataset, referred to as FDM-Bench. FDM-Bench addresses two critical areas: user query response and G-code anomaly detection, both of which are essential for improving print quality and providing effective support for FDM users. To ensure a thorough evaluation, FDM-Bench includes queries representing a broad spectrum of user expertise, from beginners to advanced researchers in FDM technology. Additionally, we generate G-codes with various types of anomalies and create multiple samples for each type by adjusting different parameters. This approach enables a comprehensive assessment of the models’ capabilities in accurately detecting these defects.

In this study, we evaluate the performance of four state-of-the-art LLMs on FDM-Bench, including two closed-source models (GPT-4o~\cite{OpenAI} and Claude 3.5 Sonnet~\cite{Anthropic}) and two open-source models (Llama-3.1-70B and Llama-3.1-405B~\cite{MetaAI}). To assess each model’s effectiveness, a panel of FDM experts evaluate responses in the user query section, focusing on accuracy, precision, and relevance. Additionally, we include multiple-choice questions (MCQs) to facilitate future evaluations by non-experts. Our results show that Llama-3.1-70B generally performs the lowest across both tasks, likely due to its smaller model size. In G-code-related tasks, closed-source models consistently outperform open-source models, while in user queries, Llama-3.1-405B achieves results comparable to the closed-source models. Our main contributions are summarized as follows:
\begin{enumerate}
    \item We introduce FDM-Bench, the first benchmark dataset for evaluating LLMs on FDM-specific tasks. FDM-Bench includes user queries across different expertise levels and G-code samples with diverse anomalies, providing a solid foundation for assessing model performance in FDM applications.
    
    \item We evaluate four state-of-the-art LLMs on FDM-Bench, including closed-source models (GPT-4o, Claude 3.5 Sonnet) and open-source models (Llama-3.1-70B, Llama-3.1-405B). Our results indicate that closed-source models generally perform better in G-code anomaly detection, while Llama-3.1-405B achieves comparable results to the closed-source models in responding to user queries.
    
    \item To support evaluations by both expert and non-expert users, FDM-Bench includes MCQs and open-ended queries. The MCQs enable efficient automated scoring, while open-ended queries allow for an in-depth assessment of model performance in supporting FDM users.
\end{enumerate}

The remainder of this paper is organized as follows. Section~\ref{sec:tasks} provides a detailed overview of tasks included in FDM-Bench. Section~\ref{sec:dataset} describes the dataset created for this study, and Section~\ref{sec:evaluation} introduces the evaluation metrics used for each task. Section~\ref{sec:results} presents a comprehensive evaluation of the LLMs on FDM-Bench. Finally, Section~\ref{sec:conclusion} concludes the paper. 
\section{Task Definition}
\label{sec:tasks}

\subsection{G-code based anomaly detection}
This task focuses on analyzing G-code to predict anomalies before the printing process begins. FDM printing is prone to defects like spaghetti (SP), under-extrusion (UE), and over-extrusion (OE), which can compromise both the quality and functionality of printed parts. These issues often arise from configuration settings or parameter values specified within the G-code. For example, UE may result from an insufficient flow rate or low nozzle temperature, whereas OE is commonly due to an excessive extrusion multiplier. Detecting and addressing these issues in the G-code before printing can help save time, materials, and costs associated with failed prints.

In this task, LLMs receive G-code from both successful and defective prints, along with information about the printer model and slicer software. The models are then asked to predict which specific anomaly is most likely to occur or if the part will print without issues.

\subsection{User queries}

This task assesses the LLMs' ability to provide accurate responses to FDM-related questions across various user experience levels. Additionally, we evaluate whether the models can identify and adapt to the user’s level of experience. To ensure both a comprehensive evaluation and efficient automatic assessment, we review the models' answers to two types of questions.

\subsubsection{Free-form questions} 

In this format, LLMs respond to questions in a free-form style. The FDM-related questions span a wide range of topics, requiring answers that are both technically and theoretically accurate. Each response must also be appropriate for the user’s experience level. Human experts and detailed qualitative metrics are essential for evaluating this type of question. This assessment allows for a thorough examination of the depth, breadth, and relevance of the LLMs’ answers.

\subsubsection{Multiple-choice questions}

MCQs with a single correct answer, or “ground truth,” are needed to enable fast and scalable evaluation of LLMs. This format allows for automatic scoring, eliminating the need for human evaluators and providing a quick assessment of how accurately each model addresses various FDM-related questions. While this setup effectively measures factual accuracy, it lacks the flexibility of free-form responses, which are essential for evaluating reasoning, relevance to user experience levels, and the exclusion of irrelevant information.

\section{Dataset Description}
\label{sec:dataset}

In this section, we introduce a dataset containing G-codes, questions, and prompts to enable a comprehensive evaluation of LLMs on the tasks outlined in the previous section. The code and dataset developed in this study are publicly available at \url{https://github.com/AhmadrezaNia/FDM-Bench}. 

\subsection{G-codes}

\begin{figure}[h]\vspace*{4pt}
\centerline{\includegraphics[scale=0.49]{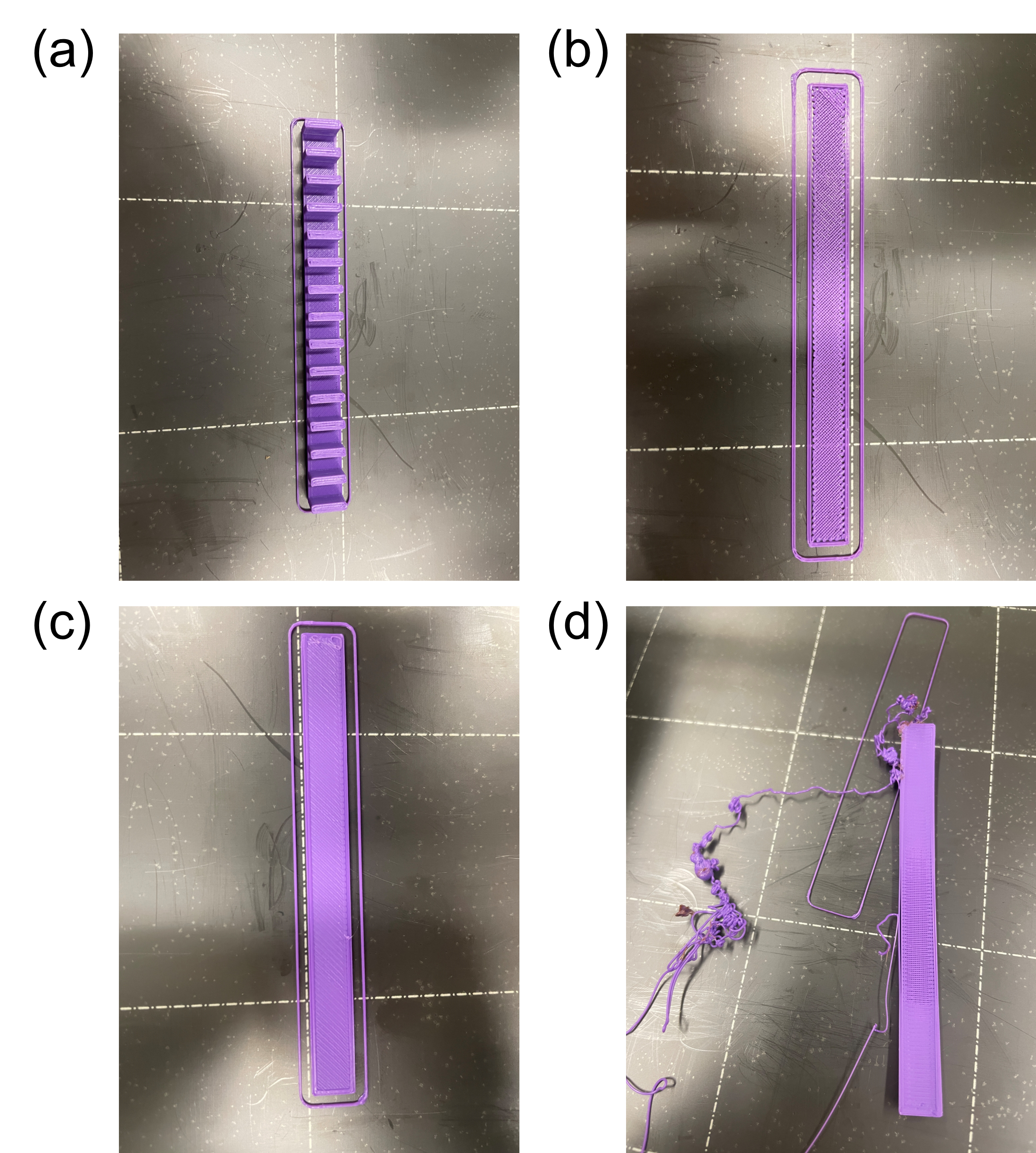}}
\caption{Printed parts illustrating different FDM quality classes: (a) ND, (b) UE, (c) OE, and (d) SP.}   
\label{fig:sample} 
\end{figure}

In this section, we describe the process of generating a G-code dataset by printing a sample bridge geometry on a Prusa 3D printer using PLA filament. By adjusting various printing parameters in Prusa Slicer software, we systematically introduce specific anomalies into the printed samples. The key parameters that are modified include bed temperature, nozzle temperature, print speed, layer height, fill angle, and flow rate (controlled through the extrusion multiplier setting). Figure~\ref{fig:sample} illustrates the four categories of printed parts in our dataset. These categories include:
\begin{itemize}
    
    \item \textbf{Non-defective (ND)} prints are created using standard, recommended parameter ranges. For example, with PLA filament, a bed temperature of 60°C and a nozzle temperature of 210°C, combined with standard flow rate settings (extrusion multiplier set to 1), yielded stable prints at typical print speeds and layer heights.
    
    \item \textbf{OE} occurs when excess filament is extruded, leading to overlapping print patterns and dimensional inaccuracies. To create over-extruded samples, we increase the extrusion multiplier above the standard setting, with values ranging from 1.3 to 1.6. While the flow rate is the primary parameter influencing OE, we also slightly adjust the bed temperature, nozzle temperature, and layer height to replicate realistic variations observed in over-extruded prints.
    
    \item \textbf{UE} occurs when insufficient filament is extruded, leading to visible gaps and weak bonding between layers. Under-extruded samples are produced by setting the extrusion multiplier below the standard value, specifically within the range of 0.6 to 0.9. As with OE, we also adjust bed temperature, nozzle temperature, and layer height.

    \item \textbf{SP} is a printing anomaly where the filament extrudes into the air rather than onto the print bed, creating a chaotic, spaghetti-like pattern. This error typically arises from two scenarios: (1) poor bed adhesion causes the print to detach and shift freely on the bed, or (2) excessive edge deformation leads to collisions between the print and the nozzle, displacing the print. In our dataset, SP is commonly induced by using low bed temperatures (35-40°C), a 90-degree fill angle (placing filament parallel to the bridge’s long axis), and high print speeds (110 mm/s).
\end{itemize}

\subsection{Questions}
In this section, we include questions relevant to a wide range of FDM users. We define three experience  levels to better evaluate the LLMs' performance in answering questions that require varying degrees of technical and theoretical knowledge:

\begin{itemize}
\item  \textbf{Beginner user} ranges from absolute beginners with no prior experience in 3D printing to early learners who are starting to understand fundamental terms and processes. This group typically requires straightforward, non-technical guidance as they become familiar with basic terminology (e.g., nozzle, filament) and settings. Their focus is primarily on understanding the essentials and troubleshooting initial challenges.

\item \textbf{Experienced user} ranges from regular operators who have been using FDM printers for several months and are familiar with standard terms, materials, and maintenance practices, to advanced operators or technicians who work with FDM printers professionally. While skilled at troubleshooting and optimizing print settings and maintenance practices, users at this level generally do not engage in research aimed at advancing the underlying FDM technology.

\item \textbf{Theoretical user} includes individuals who engage in research or theoretical work related to FDM technology. This level ranges from students and research assistants studying the FDM process to professors and faculty members conducting or supervising research. Users at this level often explore complex topics such as inter-layer bonding and material properties to advance FDM technology.
\end{itemize}

We define both free-form questions and MCQs for these three experience levels as follows:

\subsubsection{Free-form questions}
\label{sec:mcq}
The free-form questions include topics such as error diagnosis, theoretical analysis, and knowledge of slicing, hardware, and printing parameters. These questions are developed using a variety of sources, including standard references on 3D printing, textbooks, slicer software guidelines, and insights from field experts.

\subsubsection{Multiple-choice questions}
\label{}

The MCQs are concise and targeted, with each question offering five answer choices. Only one choice is correct, while the remaining four serve as distractors, designed to assess comprehension and address common misconceptions. These questions are developed using the same topics and sources as the free-form questions in Section~\ref{sec:mcq}.

\subsection{Prompts}

To evaluate the LLMs’ performance, we craft prompts tailored to the specific requirements of each task, emphasizing reasoning and factual accuracy over creative output. Accordingly, we set the temperature parameter to zero across all models. The temperature parameter, which ranges from 0 to 1, controls the randomness in a model's responses. Lower values produce more consistent outputs, while higher values introduce variability for creative responses.

In addition, we retain each model’s default settings without hyperparameter tuning. This approach ensures a fair comparison, as all models operate under identical conditions, allowing performance differences to reflect each model’s inherent capabilities rather than adjustments in parameter configurations.

To ensure clarity and consistency across tasks, prompts are structured as follows:
\begin{itemize}
    \item \textbf{Role:} The LLM is assigned a role relevant to the task.
    \item \textbf{Context:} A brief context is provided to ground the task requirements.
    \item \textbf{Task:} The LLM receives a clear, action-oriented task.
    \item \textbf{Output:} The expected output format is specified, indicating whether the response should be a single-choice answer, a free-form answer, or identify specific anomalies in the G-code.
\end{itemize}

\section{Evaluation Metrics}
\label{sec:evaluation}
This section describes the metrics used to evaluate model performance for each FDM-specific task.

\subsection{G-code anomaly detection}
We use two evaluation approaches to assess model performance in detecting anomalies from G-code.

In the first approach, models are instructed to select a deterministic label that best represents the type of anomaly present in the G-code. The labels include specific anomaly types, such as SP, OE, UE, or an ND label when no anomalies are detected. For this evaluation, we measure model performance using accuracy, reflecting how often the model correctly identifies the ground truth label for each G-code instance.

In the second approach, LLMs analyze the G-code and assign a probability to each possible label, expressed as a percentage. Here, we focus on the probability assigned to the actual ground truth label as the primary performance measure. This approach offers insights into the model’s confidence in its predictions and its capability to differentiate accurately among various potential anomalies.
 
\subsection{User queries}
\subsubsection{Free-form questions} 
A panel of FDM experts with relevant research expertise evaluates LLM responses to free-form questions. To assess response quality, each evaluator scores them on a scale of 1 to 5 for each of the following three criteria, with 5 indicating the highest performance and 1 the lowest:

\begin{itemize}
    \item \textbf{Accuracy} assesses the factual accuracy of the response, focusing on alignment with the essential content required to answer the question. Higher scores indicate that the response comprehensively includes all essential correct information while avoiding any incorrect information.

    \item \textbf{Precision} measures the focus of the response by evaluating how free it is from unnecessary or extraneous content. Higher precision scores indicate concise responses that contain only the information directly relevant to the question.

    \item \textbf{Relevance to experience level} evaluates how effectively the response aligns with the specified user expertise level. Responses that provide appropriately detailed information for the intended user's level receive higher scores, while responses that are too simplistic or overly complex for the specified skill level are rated lower.
\end{itemize}

\subsubsection{Multiple-choice questions}
We automatically evaluate the LLMs’ accuracy in selecting correct answers to each MCQ. Accuracy is determined as the percentage of questions answered correctly by comparing responses with the ground truth. This metric directly measures the LLM's effectiveness in retrieving FDM-related information.

\section{Results and Discussions}
\label{sec:results}

\subsection{Models}
In this study, we evaluate the performance of four state-of-the-art LLMs on FDM-specific tasks. The selected models include both closed-source and open-source options, allowing for comparative analysis across architectures, model sizes, and accessibility. Each model undergoes identical tasks to assess its capabilities in FDM anomaly detection and responding to user queries.

The closed-source LLMs include:
\begin{itemize}
    \item \textbf{GPT-4o}: Developed by OpenAI, GPT-4o is an advanced iteration of the Generative Pre-trained Transformer series, released in May 2024. While the exact parameter count remains undisclosed, GPT-4o supports a maximum input context window of 128,000 tokens and can generate up to 16,384 tokens in a single response.
    
    \item \textbf{Claude 3.5 Sonnet}: Anthropic's Claude 3.5 Sonnet, released in June 2024, is designed to excel in natural language understanding and generation. While the exact parameter count is undisclosed, it supports a maximum input context window of 200,000 tokens and can produce up to 4,096 tokens in length. Hereafter, any reference to Claude refers specifically to this model.
\end{itemize}

The open-source LLMs include:
\begin{itemize}
    \item \textbf{Llama-3.1-70B}: Released by Meta in July 2024, this model comprises 70 billion parameters and supports an input context window of 128,000 tokens, allowing for the effective processing of extensive textual data.
    
    \item \textbf{Llama-3.1-405B}: Introduced by Meta in July 2024, this larger model includes 405 billion parameters. Designed for handling complex tasks with high accuracy, it also supports an input context window of 128,000 tokens.
\end{itemize}

\subsection{Performance comparison in anomaly detection}
This section compares the anomaly detection performance of four LLM models through both deterministic labeling and the probabilistic scoring method.

\subsubsection{Deterministic labeling}

\begin{figure}[h]\vspace*{4pt}
\centerline{\includegraphics[scale=0.39]{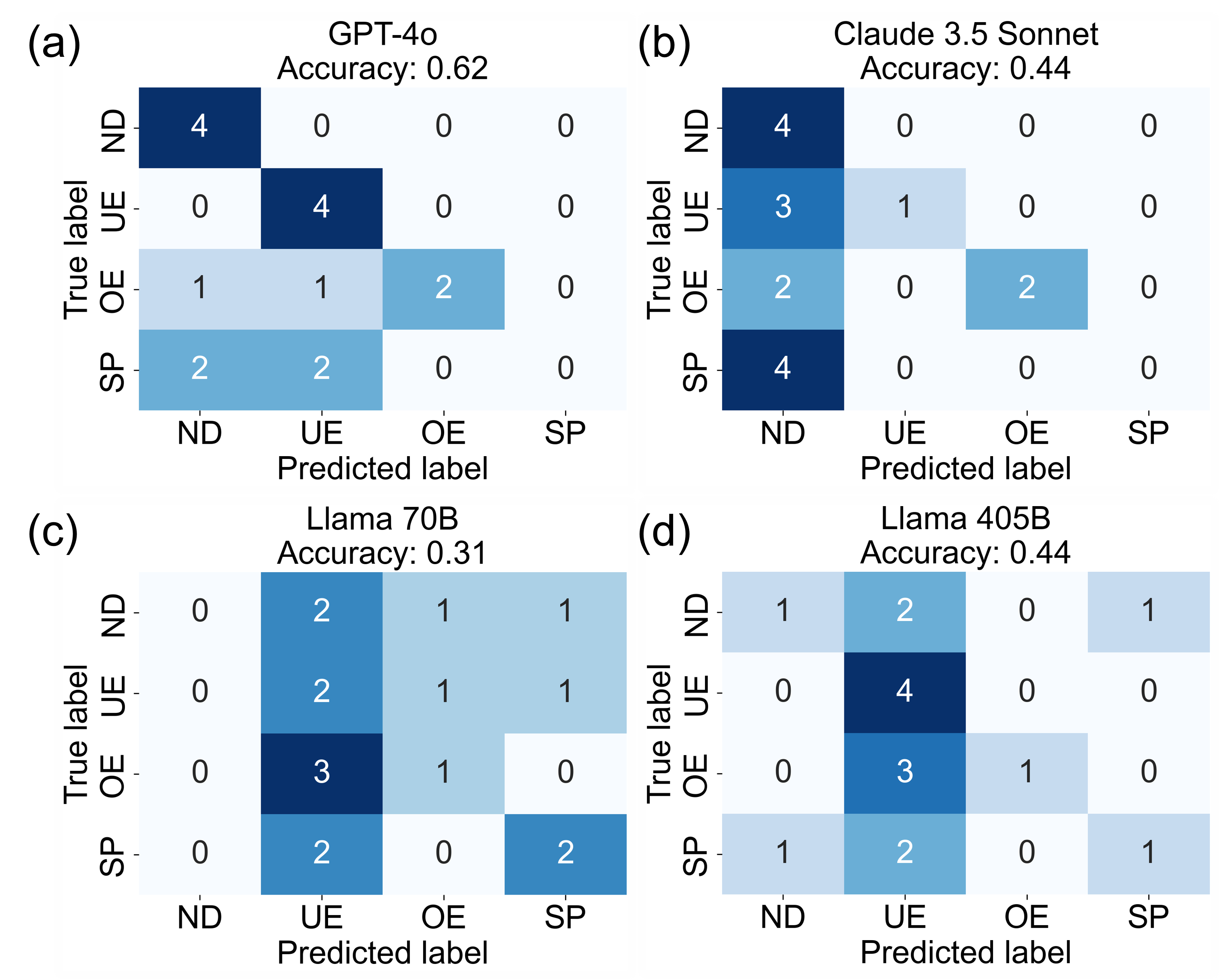}}
\caption{Confusion matrices for G-code anomaly detection across four LLM models: (a) GPT-4o, (b) Claude 3.5 Sonnet, (c) Llama 70B, and (d) Llama 405B.}   
\label{fig:confusion_all} 
\end{figure}

Figure~\ref{fig:confusion_all} presents the confusion matrices for each LLM evaluated using the deterministic approach. Each model is evaluated with four data samples per class across the four categories: SP, UE, OE, and ND. GPT-4o achieves the highest accuracy, correctly identifying the ground truth anomaly label in 62\% of cases. Llama 405B and Claude followed, each achieving an accuracy of 44\%. Llama 70B demonstrated the lowest accuracy at 31\%, slightly above random guessing (25\%). These results suggest that all models, including Llama 70B, have some capacity for interpreting G-codes.

The confusion matrices reveal specific misclassification patterns, with each model showing a tendency to favor certain labels. For instance, Claude and Llama 405B primarily assigned the ND and UE labels, respectively. The other two models show a similar, though less pronounced, trend. Notably, Llama 70B and Llama 405B exhibit similar preferences in label assignment, potentially due to both models being developed by Meta AI~\cite{MetaAI} and likely trained on similar datasets.

Additionally, both open-source Llama models demonstrate a more cautious approach to anomaly detection and rarely label parts as ND. In contrast, the closed-source models (GPT-4o and Claude) are less conservative and tend to label more parts as ND than they actually are. This difference in labeling tendencies might reflect different approaches to anomaly detection across these models.

\subsubsection{Probabilistic scoring}

In the probabilistic scoring approach, we utilize the same set of G-codes used in the deterministic method. However, instead of requesting a single label, we instruct each LLM to assign a likelihood to each label in percentage terms, with probabilities summing to 100\% for each G-code sample.

\begin{figure}[h]\vspace*{4pt}
\centerline{\includegraphics[scale=0.56]{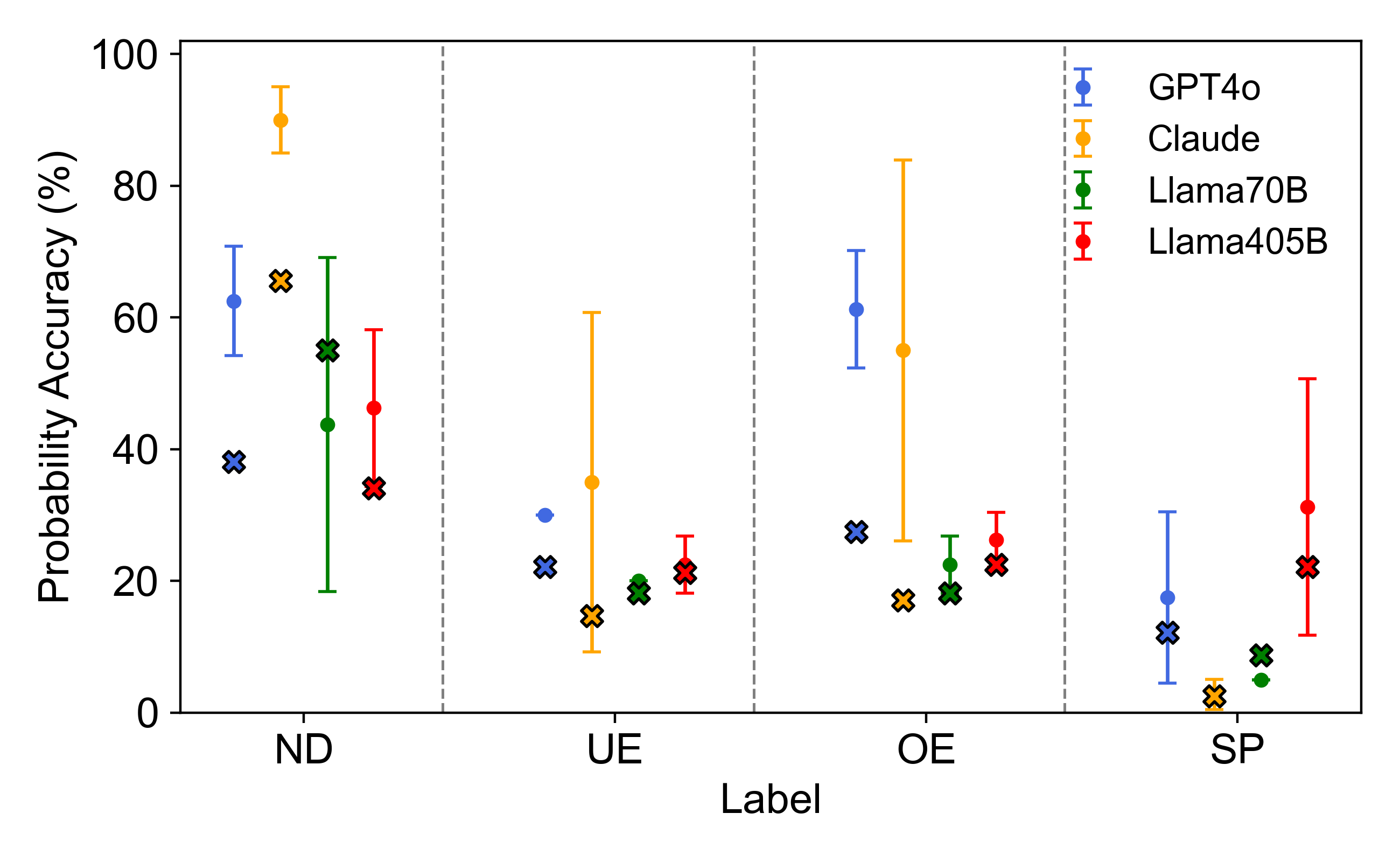}}
\caption{ Average probability assigned by each LLM model to the correct label for each G-code class. Crosses indicate the average probability assigned to each label across all samples.}   
\label{fig:detect-prob} 
\end{figure}

Each LLM produces four probability values per sample, corresponding to the likelihood of each class. Figure~\ref{fig:detect-prob} presents the models' performance by showing the average probability assigned to the correct label for each class. We compute the average probability for the correct label across the four samples per class, with the standard deviation reflecting the consistency of each model’s performance across these samples.

Additionally, to account for potential biases seen in the deterministic approach, where certain models favored specific labels, we calculate the average probability assigned to each label across all samples for each model. These average probabilities are marked with crosses in Figure~\ref{fig:detect-prob}. Comparing the average correct label probability with overall label preferences provides insights into whether each model effectively differentiates among distinct G-code patterns or only favors certain labels regardless of input.

As shown in Figure~\ref{fig:detect-prob}, the probabilistic scoring results show that GPT-4o, Claude, and Llama 405B tend to assign higher probabilities to the correct labels (indicated by circles) compared to their average probability assignment across all labels (indicated by crosses). This pattern suggests that these models effectively analyze G-codes, extract relevant information, and detect anomalies accurately. In contrast, Llama 70B does not show a clear distinction between correct and incorrect labels, indicating a limited ability to identify anomalies accurately. This weaker performance may be due to the smaller size of Llama 70B compared to the other three models. Additionally, this task involves a near-maximum input token size, potentially limiting Llama 70B’s capacity to fully interpret the G-code data.

A closer examination of Claude’s performance reveals that although it detects ND, UE, and OE G-codes with fairly high accuracy, it consistently assigns low probabilities to the SP anomaly, indicating a potential blind spot for this specific defect. This limitation may reflect a gap in Claude’s training data or an inherent model bias. Additionally, the high variability in Claude’s results, reflected by a larger standard deviation, suggests inconsistent predictions across different samples. This inconsistency could impact Claude’s reliability in real-world applications, where stable performance is essential.

An analysis of GPT-4o’s results indicates that its labeling variation remains within a reasonable range and assigns consistently higher probabilities to correct labels than the average probability for each label. This consistency aligns with the deterministic labeling results, where GPT-4o outperforms the other models, highlighting its potential as a more robust choice for G-code anomaly detection tasks.


\subsection{Performance comparison in answering free-form questions}

\begin{figure}[h]\vspace*{4pt}
\centerline{\includegraphics[scale=0.56]{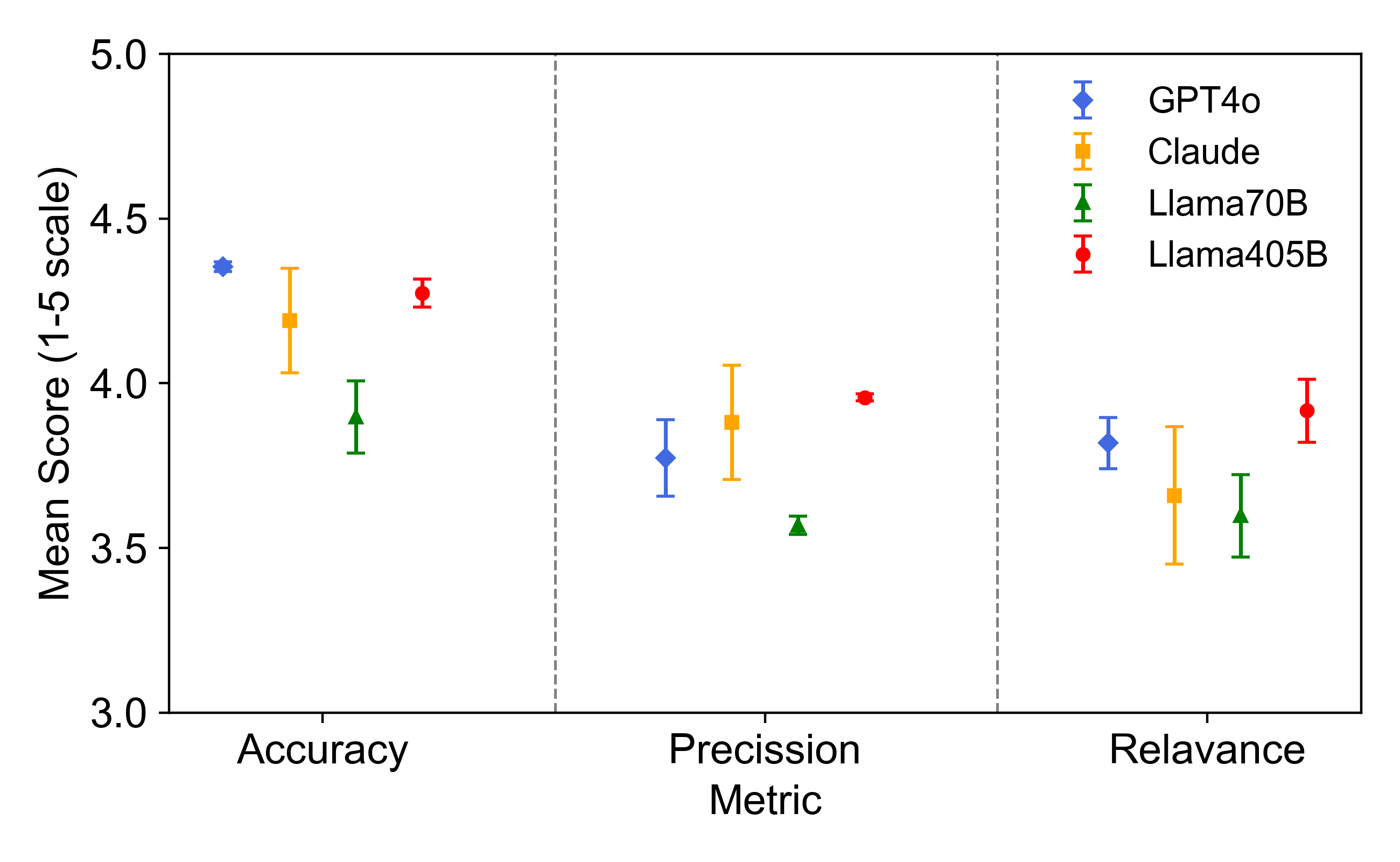}}
\caption{Average scores (1–5 scale) for each LLM model on free-form user query responses across three metrics: accuracy, precision, and relevance.}   
\label{fig:free-overal} 
\end{figure}

This section investigates the performance of LLMs in answering free-form user queries across beginner, experienced, and theoretical FDM users. Fourteen evaluators score these responses across three key metrics: accuracy, precision, and relevance to user experience level, on a scale from 1 to 5. Figure~\ref{fig:free-overal} presents each model’s average performance on these metrics, with standard deviation bars indicating the variability across different expertise levels.

As shown in Figure~\ref{fig:free-overal}, GPT-4o achieves the highest accuracy among the models, with minimal variation, suggesting consistent performance in accurately addressing questions across all user levels. Following GPT-4o, Llama 405B ranks second in accuracy, with a small margin difference, indicating strong performance. In terms of relevance and precision, Llama 405B outperforms the other models, highlighting its ability to provide responses that are concise and appropriately tailored to each user’s expertise level. In contrast, Llama 70B demonstrates the lowest performance across all three metrics. Notably, Claude exhibits the largest variability in its scores, indicating less consistent responses across different user levels.

These findings suggest that while GPT-4o excels in delivering accurate answers, Llama 405B is more effective in ensuring relevance and precision, albeit with a slight trade-off in accuracy.

\subsection{Performance comparison in answering multiple-choice questions}

\begin{figure}[h]\vspace*{4pt}
\centerline{\includegraphics[scale=0.60]{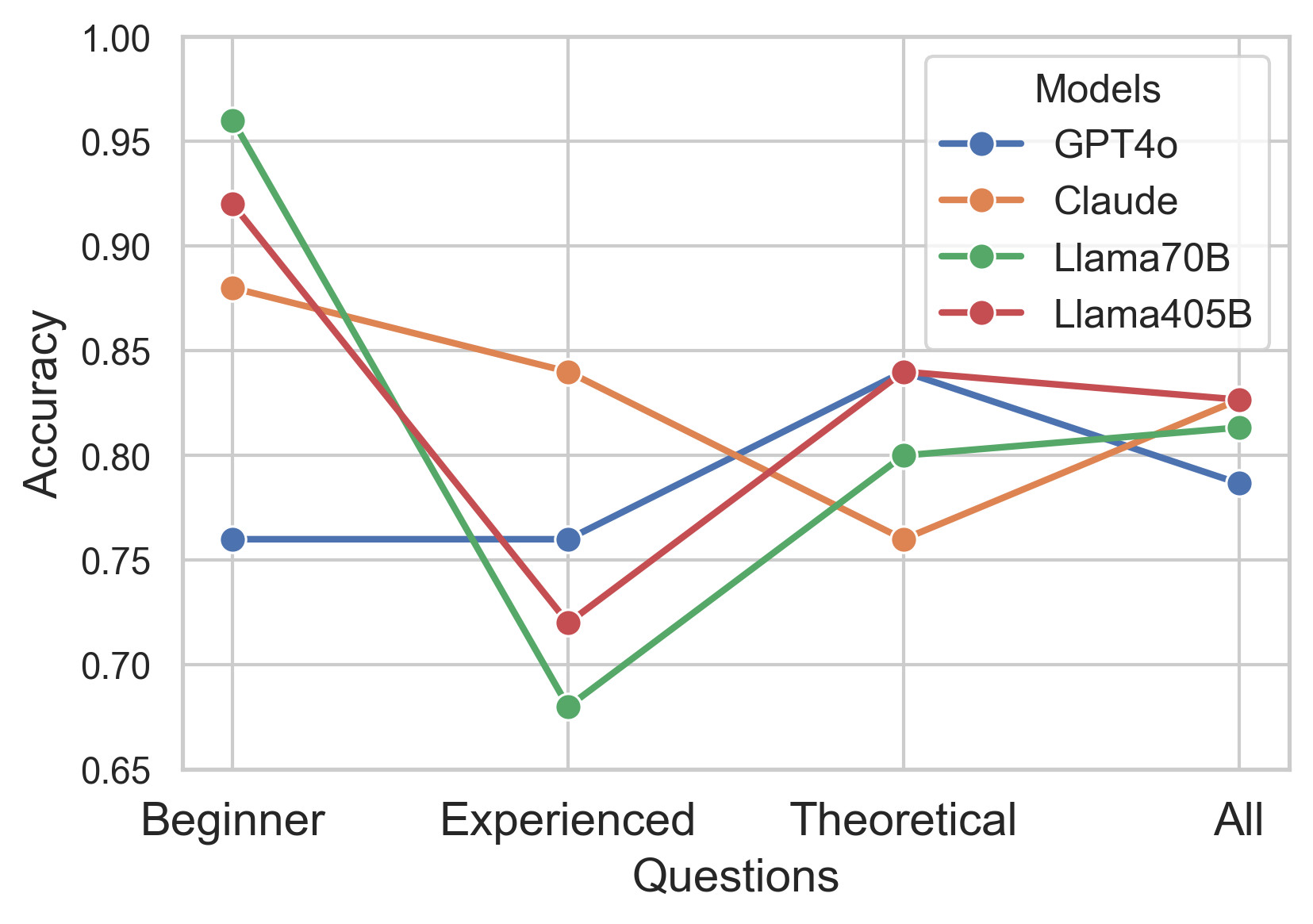}}
\caption{Accuracy comparison of LLM models in answering multiple-choice questions across different user expertise levels.}   
\label{fig:Multi} 
\end{figure}

This section compares the performance of the LLMs in answering MCQs across three user expertise levels: beginner, experienced, and theoretical. Figure~\ref{fig:Multi} shows the accuracy of each model across these categories.

Overall accuracy, calculated as the average across all question levels, shows that the models perform similarly, with Claude and Llama 405B achieving slightly higher accuracy than the others. However, examining accuracy across user levels reveals clearer differences among the models. Notably, the top-performing model varies by user level. For instance, Llama 70B achieves the highest accuracy for beginner-level questions, while Claude performs best at the experienced level. However, neither ranks highest for theoretical questions.

Additionally, individual model performance shifts significantly across expertise levels. For example, while Llama 70B performs best on beginner questions, it shows the lowest accuracy at the experienced level, highlighting differences in model adaptability across expertise levels. As expected, most models perform best on beginner questions, likely due to their simpler content. Interestingly, overall accuracy is higher for theoretical questions than for experienced questions, possibly reflecting the nature of the training data used for these models.

These findings suggest that LLM effectiveness varies significantly with question complexity and technical demands. This indicates that some models may be better suited to specific user expertise levels.


\section{Conclusion and Future Work}
\label{sec:conclusion}

In this study, we introduce FDM-Bench, the first comprehensive benchmark designed to evaluate LLMs on tasks specific to FDM. Two closed-source models (GPT-4 and Claude 3.5 Sonnet) and two open-source models (Llama-3.1-70B and Llama-3.1-405B) are evaluated on user queries and G-code samples. Our findings highlight the unique strengths each model offers for different FDM-related tasks. Although closed-source models generally achieve higher accuracy in G-code anomaly detection, Llama-3.1-405B also demonstrates promising results, particularly in its cautious approach to defect identification. Unlike the closed-source models, which more confidently label some anomalous G-codes as ND, Llama-3.1-405B shows a more conservative and potentially advantageous strategy in quality control contexts. Additionally, Llama-3.1-405B slightly outperforms the closed-source models in responding to user queries, demonstrating that open-source models can achieve performance comparable to state-of-the-art closed-source options.

These findings underscore the potential for further enhancement of open-source models through fine-tuning~\cite{Hu} or retrieval-augmented methods~\cite{Lewis} to achieve better performance in FDM-specific tasks. By establishing a standardized evaluation framework, FDM-Bench supports consistent benchmarking of these improved LLMs, thus contributing to more effective LLM applications in AM.

Future expansions of FDM-Bench can focus on broadening its applicability across a wider range of AM tasks. This will include evaluations of additional AM technologies such as selective laser sintering, stereolithography, and metal AM~\cite{Chandrasekhar}. Another promising direction is the inclusion of image-based tasks that utilize large vision-language models (LVLMs)~\cite{Li, Liu, Zhu} to improve visual defect detection and enable the prediction of anomalies arising from external causes, rather than those solely related to printing parameters and settings~\cite{Jadhav}. Together, these expansions are expected to establish a more comprehensive framework for evaluating LVLM performance across a broad spectrum of AM applications and settings.

Additionally, given the sensitivity of these models to prompt structure~\cite{Arora, Bhargava}, incorporating advanced prompting techniques, such as few-shot prompting~\cite{Brown}, chain-of-thought~\cite{Wei}, and tree-of-thought~\cite{Yao} prompting, may offer deeper insights into their reasoning processes. These techniques could enhance model accuracy and consistency in both reasoning and calculations within AM contexts.

\section*{Acknowledgements}

This work was supported by the National Science Foundation (NSF) under grants 2126246, 2434383, and 2434385, and the U.S. Department of Agriculture (USDA) under grant AG Sub UCDavisA21-0845-S003 S.

\clearpage\onecolumn

\end{document}